\documentclass[letterpaper, 10 pt, conference]{ieeeconf}  

\IEEEoverridecommandlockouts                              

\usepackage{mathtools}
\usepackage{amsmath}
\usepackage{pdfpages}
\usepackage{cite}
\usepackage{hyperref}
\usepackage[normalem]{ulem}

\usepackage[ruled,vlined]{algorithm2e}

\DeclarePairedDelimiter\floor{\lfloor}{\rfloor}
\DeclareMathOperator*{\argmax}{argmax} 

\title{\LARGE \bf
Understanding Contexts Inside Robot and Human Manipulation Tasks through a Vision-Language Model and Ontology System in a Video Stream
}

\author{Chen Jiang$^{\dagger}$, Masood Dehghan$^{\dagger}$ and Martin Jagersand$^{\dagger}$
\thanks{$^{\dagger}$Authors are with Department of Computing Science,
        University of Alberta, Edmonton AB., Canada, T6G 2E8.
        { 
           \tt\small \{cjiang2, masood1, mj7\}@ualberta.ca
        }
        }%
}

\begin{document}

\maketitle
\thispagestyle{empty}
\pagestyle{empty}

\begin{abstract}
Manipulation tasks in daily life, such as pouring water, unfold intentionally under specialized manipulation contexts. Being able to process contextual knowledge in these Activities of Daily Living (ADLs) over time can help us understand manipulation intentions, which are essential for an intelligent robot to transition smoothly between various manipulation actions. In this paper, to model the intended concepts of manipulation, we present a vision dataset under a strictly constrained knowledge domain for both robot and human manipulations, where manipulation concepts and relations are stored by an ontology system in a taxonomic manner. Furthermore, we propose a scheme to generate a combination of visual attentions and an evolving knowledge graph filled with commonsense knowledge. Our scheme works with real-world camera streams and fuses an attention-based Vision-Language model with the ontology system. The experimental results demonstrate that the proposed scheme can successfully represent the evolution of an intended object manipulation procedure for both robots and humans. The proposed scheme allows the robot to mimic human-like intentional behaviors by watching real-time videos. We aim to develop this scheme further for real-world robot intelligence in Human-Robot Interaction.
\end{abstract}

\section{Introduction}

Recent advances in fusing computer vision with linguistic knowledge enables researchers to model human-like commonsense knowledge using semantic contexts for intelligent robots. Studies both in robot vision and Natural Language Processing (NLP) \cite{yang2015robot, yang2018learning, nguyen2018translating, zhang2019robot, nguyen2019v2cnet} provide promising tools for robots to better understand human tasks and assist humans in their daily life. Still, intelligent robots are far from perfect. Intelligent robots face challenges in: (1) interpreting sensor inputs of vision and force contact interactions through modeling and learning from daily life knowledge; and (2) performing intelligent actions that take into account the surrounding physical environment as well as human intention. Various studies \cite{yang2014manipulation, Zhang2018AMC, Welschehold2019CombinedTA, fox2019multi, Lee2019AMH, takayanagi2019hierarchical, French2019LearningBT, Colledanchise2016TowardsBR, yang2015robot, yang2018learning, zhang2019robot} have structured and planned manipulation actions and activities for robotics in ways similar to human thinking, however it is still challenging to extract contextual knowledge directly from daily life. 

Understanding contexts semantically is important for robotics because humans express intention during the process of action execution. For example, in the pouring manipulation task, the context involves a sequence of actions executed over time, including, grasping an object that contains liquid, pouring the liquid into an empty container, and releasing the currently held pouring object. From studies in action recognition \cite{Swathikiran2018attention,lu2019deep, li2019manipulation}, we know that the execution of manipulation tasks requires hand-eye coordination, and humans intentionally focus their visual attention onto relevant regions of objects. Commonsense knowledge in a pouring task can be summarized by humans as constraint-like rules (e.g. "grasping an object full of liquid is needed before pouring", "destination of pouring should usually be an empty container that I can grasp safely", etc.). There is no doubt that attention and commonsense knowledge on-scene can serve as strong prior knowledge to logically deduce human intention. To plan for future manipulation actions, intelligent robots will need to infer from contextual knowledge in real-time. Therefore, it is essential to develop techniques to semantically process visual information and interpret manipulation actions on-scene for both robot and human. 

\begin{figure}[t!]
\centering
    \includegraphics[scale=0.34]{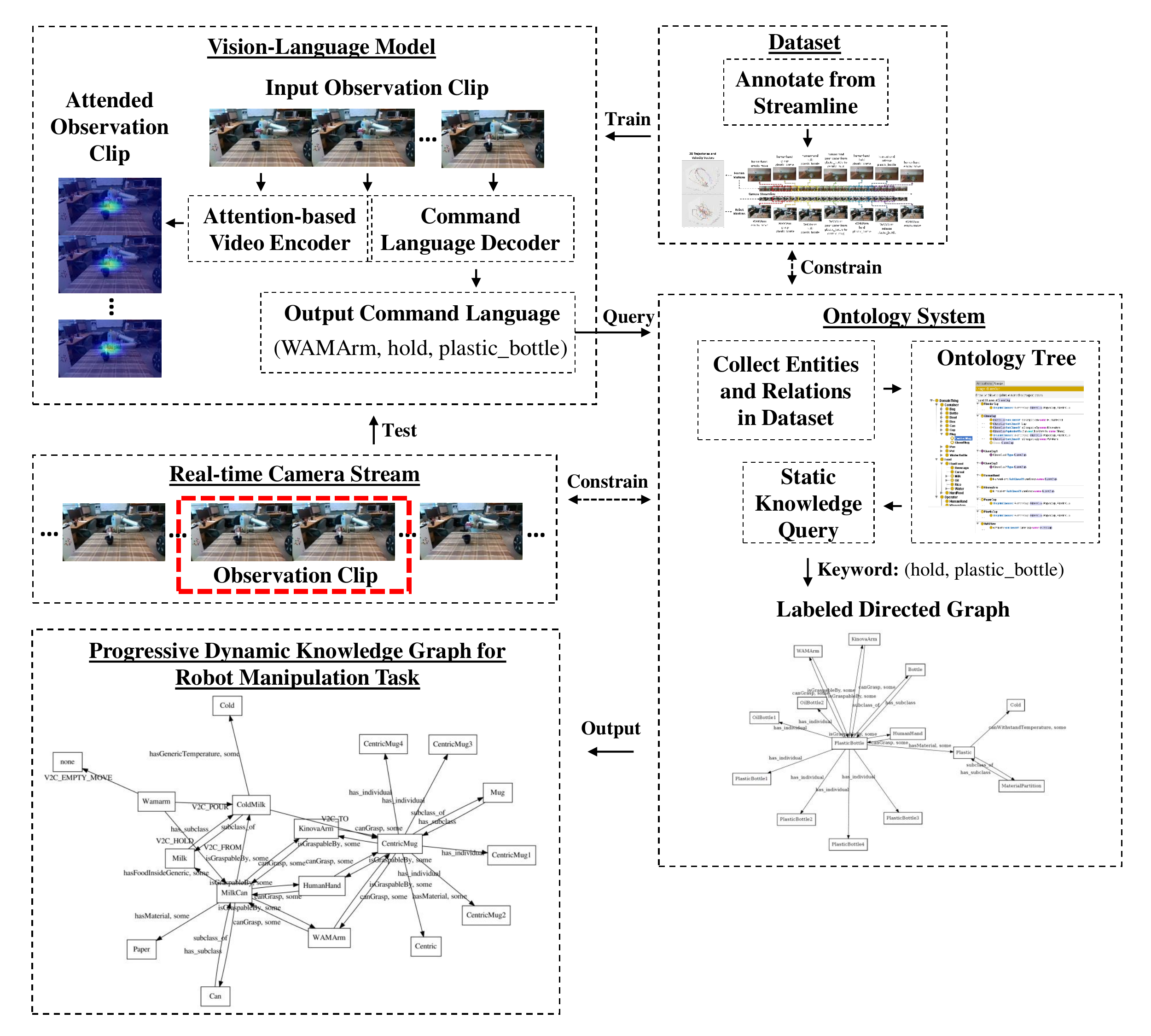}
    \caption{Overview of our scheme. Under our ontology system for commonsense knowledge, our proposed scheme is able to utilize a Vision-Language model and an ontology system to extract spatial attentions consecutively over a camera stream, and interpret a manipulation task into an evolving dynamic knowledge graph filled with concepts of manipulation. }
     \label{fig:intro}
\end{figure}
In this paper, we propose a scheme to explore the fundamental problem of perceiving and interpreting on-scene dynamic knowledge and commonsense knowledge over time for both robots and humans using an attention-based Vision-Language model and an ontology system. The purpose is to represent the evolution of the intended manipulation procedure, and generate visual attention tracks and dynamic knowledge graphs that can be used as inputs for robotic manipulation decisions. Figure \ref{fig:intro} shows the logic flow of this scheme. Our contributions can be summarized as follows:

\begin{itemize}

\item We present a scheme to capture manipulation concepts into a time-independent knowledge domain. An ontology system is constructed to store objects and relations in a taxonomic structure, which serves as our commonsense knowledge over a particular domain of manipulation tasks. 

\item We collect a benchmark dataset of RGB-D videos from manipulation tasks performed by both robots and humans. Under our domain knowledge constraints, we manually annotate ground truth object and action information to the video frames.

\item We present a Vision-Language model based on the popular sequence-to-sequence structure \cite{Sutskever2014SequenceTS} with spatial attention mechanism to caption manipulation knowledge for video streams. The Vision-Language model is able to implicitly learn spatial attention on the salient regions corresponding to the manipulation actions and activities at hand. 
\item We combine the Vision-Language model with the ontology system, allowing the model to semantically interpret the evolution of the manipulation task into a linguistic dynamic knowledge graph filled with commonsense knowledge.
 
\end{itemize}

The rest of the paper is organized as follows: section II summarizes the recent advances in intelligent robotics with contextual knowledge; section III discusses the formulation and scheme for our contextual knowledge modeling; experimental details and analysis are conducted in section IV; and we draw the final conclusion and summarize possible future works in section V. Our code and collected dataset are publicly available at: \url{https://github.com/zonetrooper32/robot_semantics}.

\section{Related Work}
\subsection{Manipulation Knowledge in Robotics}
Various methods have studied the effect of introducing contextual knowledge for robotic behaviors. Multiple studies \cite{aditya2015visual, ye2017can, aditya2018image, beetz2018know, aditya2019integrating, Ziaeetabar2018PredictionOM} have discussed generic schemes to represent collective commonsense knowledge on manipulation objects and relations for scene understanding and commonsense reasoning. The evolution of manipulation tasks is another wildly studied aspect that directly utilizes manipulation knowledge in robotics. Task evolution can be represented by structures such as semantic trees \cite{yang2014manipulation, Zhang2018AMC, Welschehold2019CombinedTA, fox2019multi}, state transition graphs \cite{Lee2019AMH, takayanagi2019hierarchical} or behavior trees \cite{French2019LearningBT, Colledanchise2016TowardsBR}. However, most of those evolution representations rely heavily on human annotations, and few of the aforementioned studies have discussed how to automatically acquire evolution representations on-scene. In our work, initiated from human knowledge in manipulation contexts, we utilize a Vision-Language model to automatically interpret task evolution and robotic actions semantically. 

\subsection{Vision and Language in Robotics}
Originating from action recognition and video captioning, there have been a number of studies on introducing language in combination with vision to learn semantic actions for robotic uses. Nguyen et al. \cite{nguyen2018translating, nguyen2019v2cnet} proposed to caption human actions into command sentences, which can be used to control robotic actions. Similar works can be observed to improve the capabilities of Vision-Language models under robotic settings for problems like Human-Robot Interaction \cite{Hatori2017InteractivelyPR, Shridhar2018InteractiveVG, Thomason2019ImprovingGN}, action learning and planning \cite{yang2015robot, yang2018learning, zhang2019robot, Paxton2019ProspectionIP, Zhang2019AnOA}, etc. However, the evaluation of these methods usually involves: (1) sampling of a small fixed number of frames, which is not suitable when intermediate feedback is continuously requested in a real-time video stream; or (2) heavy reliance on object detection, which is only weakly associated to manipulation contexts. We highlight these points in our scheme with a Vision-Language model under a more realistic semantic context using video streams. 

\section{Robot Semantics}

In this section, we introduce our main framework to semantically model contextual knowledge in both robot and human manipulation tasks. We first propose a general scheme to model the timing constraints typically found in robot vision, then present our scheme to collect, construct and model contextual knowledge for manipulation tasks in general. 

\subsection{Sampling from Streams} 

\subsubsection{Stream, Video and Dataset} A camera stream $CS_{T_0}$, from start time $T_0$ to (potentially) infinity, observes a scene of a human or robot performing a sequence of actions and produces image frames $I_{T_k}$. We define a video of length $j - i + 1$ in the form $V_{T_i...T_j} = \{I_{T_i}, I_{T_{i+1}}, ..., I_{T_j}\}$, initiating from start time $T_i$ to end time $T_j$. Each video shows a full demonstration over a sequence of actions. A dataset $D = \{V_1, V_2, ..., V_N\}$ is defined as a collection of $N$ videos from multiple camera streams of different time periods with annotations. 

\begin{figure}[ht]
\centering
    \includegraphics[scale=0.6]{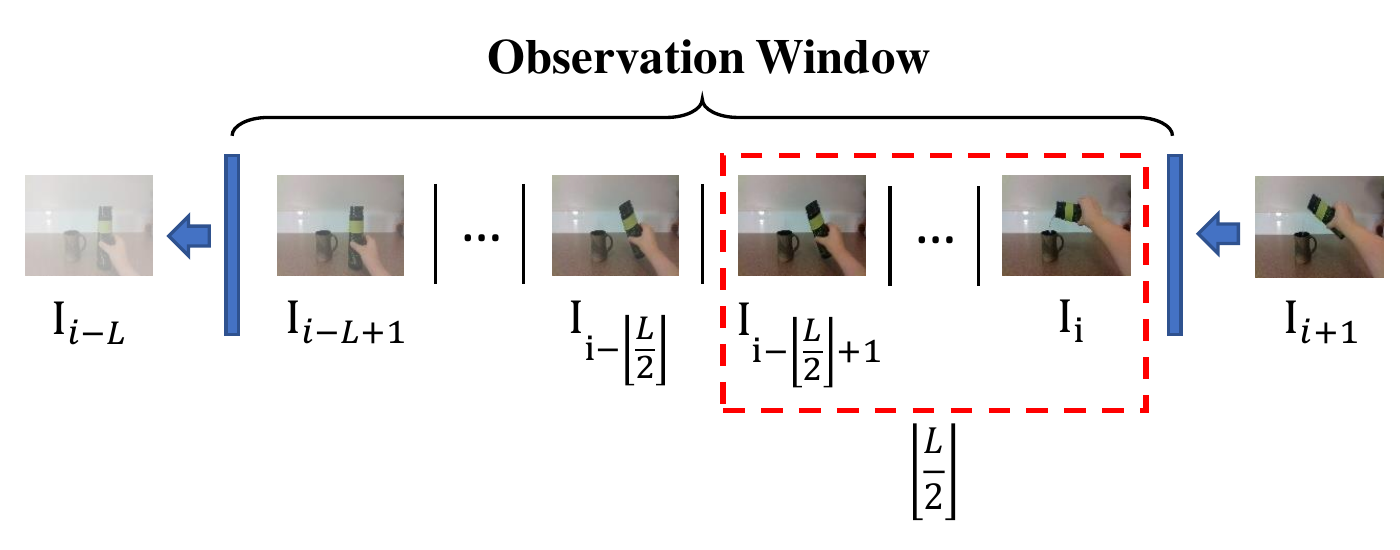}
    \caption{Representing camera vision as a stream of data, where new data is queued while outdated data is flushed. A queue of observations is maintained and stream signals for inference when $\floor{\frac{L}{2}}$ of the observation is queued with new information.}
     \label{fig:streamline}
\end{figure}

\begin{figure*}[!htp]
\centering
\includegraphics[scale=0.3]{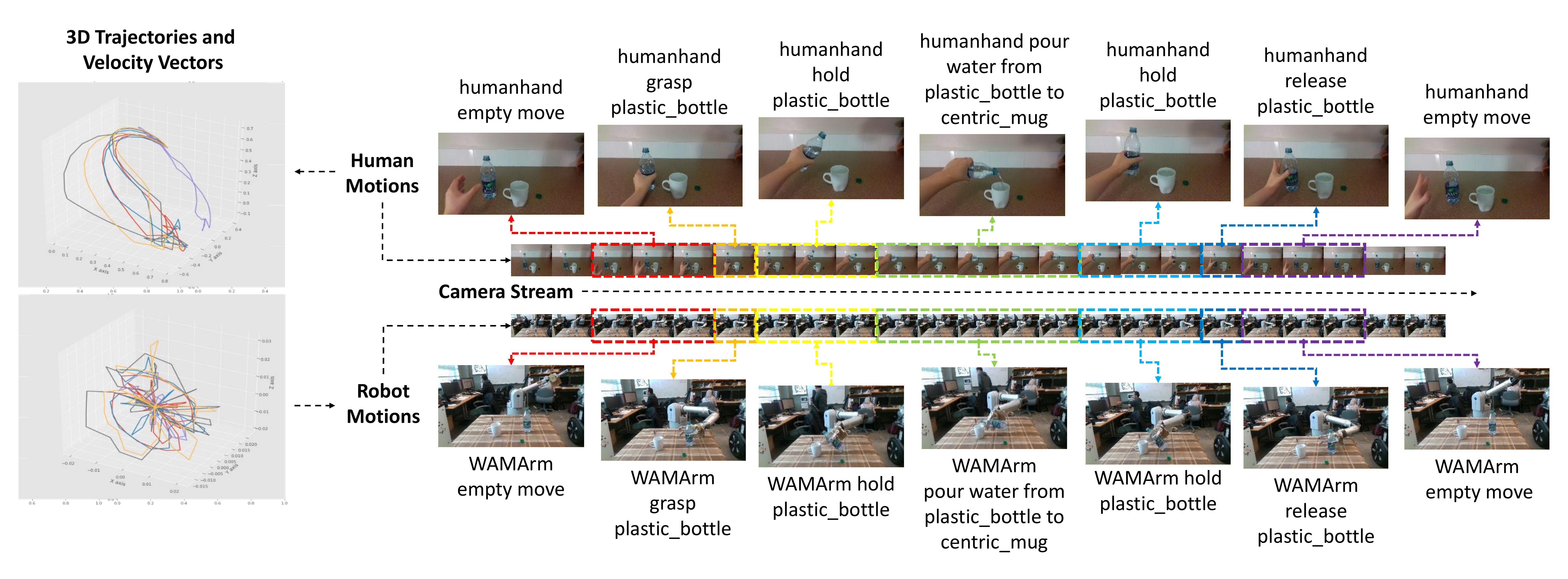}
\caption{Samples from our Robot Semantics dataset. Initiated from human motions, manipulation videos are collected for both human and WAM robot arm, shown side-by-side with action and object annotations. Left: Ten of the 3D trajectories and the velocity vectors from the robot motions used in video collection. }
\label{fig:rs_rgbd_dataset}
\end{figure*}

\subsubsection{Observation} We assume weak to no long-term time dependencies, i.e. an observation is "most dependent and trust-able" within a time window. We further define this intermediate observation as a small clip of frames $C_{t_k...t_{k+L}} = \{I_{t_k}, I_{t_{k+1}}, ..., I_{t_{k+L}}\}$, initiating from the start time $t_k$ and persisting for a time length $L$. Figure \ref{fig:streamline} presents a simple scheme to sample a series of overlapping clip observations from a camera stream. The goal is to always maintain a queue of maximum $L$ length serving as our observation window, while streams of image frames continuously arrive into the observation window. A clip is collected when: (1) the observation window is filled for the first time; or (2) $\floor{\frac{L}{2}}$ of the observation window is flushed with newer images. Hence, the camera stream  $CS_{T_0}$ is treated as an overlapping sequential composition of several $L$ length clips.

\subsection{Robot Semantics Dataset}

While many datasets exist for manipulation tasks and human intentions \cite{mahdisoltani2018effectiveness, nguyen2018translating, damen2018scaling, li2018eye}, few span both robot and human manipulation tasks. We collect 720p videos over a set of manipulation tasks using an Intel RealSense D435i Camera. Figure \ref{fig:rs_rgbd_dataset} presents an overview over our benchmark data with command language annotations and robot way-point trajectories. Our Robot Semantics Dataset consists of RGB-D videos, where each video demonstrates a complete particular manipulation task such as "pouring water to the cup". Objects are placed at random locations, and a manipulator (robot or human) executes a sequence of motions to complete the full manipulation task. During the process, our command language, actions and object-types, are annotated frame-by-frame. We focus on manipulation actions such as "grasp", "release", "pour", "hold", and "intent". We included two types of manipulators in our dataset: human subjects and a Barrett Whole Arm Manipulator (WAM) robot.

\subsubsection{Human} 
The camera was setup with an egocentric view in a kitchen environment. Human subjects were asked to use their left or right hand to perform a series of actions for a complete manipulation task. For manipulation tasks performed by a human, there are 94 videos - 42,681 images in total. 

\subsubsection{WAM}
A Barrett WAM robot was used to perform the same set of manipulation tasks as the human subjects. The camera was setup to view the majority of the WAM and a kitchen table top with objects on it. We used the experimental protocol originating from IVOS benchmark of Siam et al \cite{siam2019video}; a human operator guided the WAM to reach the target and perform the intended manipulation actions. Robotic way-point trajectories, in the form of quaternions over the 7 joints poses, were recorded during the kinesthetic teaching. The WAM robot then executed the manipulation actions by following the recorded way-point trajectories. For the WAM robot arm, there are 46 videos - 69,368 images and 43 recorded trajectories in total. 

\subsection{Domain Knowledge Ontology}

An ontology is a well known way to store machine-interpretable definitions of concepts in a static knowledge domain. Here, we construct an ontology to store and query for the set of explicitly defined commonsense knowledge over the concepts of manipulation. Given a set of linguistic entities $E = \{e_1, e_2, ..., e_n\}$, including the objects and manipulators presented in our dataset, for any two entities $e_i, e_j \in E$, we impose a set of binary logical constraints $LC$ in a taxonomic and relational structure using a linguistic relation $a_k \in A$:
$$
e_i \xrightarrow{\text{$a_k$, $r$}} e_j \in LC \eqno{(1)}
$$
\noindent where $A = \{a_1, a_2, ..., a_m\}$ is the set of relations, $r$ represents restrictions (Quantifier, Cardinality, and hasValue). $E + A$ represents the complete linguistic vocabulary over the entire manipulation domain knowledge. In general, relations can originate from: (1) hierarchical definitions, for example, \textit{``$\forall_{(GlassCup, Cup)}$ isA(GlassCup, Cup)"}, \textit{``$\forall_{(PlasticBottle, HardBottle)}$ disjoint(PlasticBottle, HardBottle)"}; (2) action relation between any two entities, for example, \textit{``$\forall_{WAM}$ canPour some ColdMilk"} and \textit{``$\forall _{PlasticBottle}$ isGraspableBy some HumanHand"}; and (3) attributes or properties of any entity, for example, \textit{``$\forall_{ColdMilk}$ hasTemperature some Cold"}, \textit{``$\forall_{GlassCup}$ canHold some HotWater"}, etc. Entities are stored as classes with individuals, while relations are stored as binary object properties with restrictions. 

\subsection{Defining Dynamic Knowledge}
To interpret on-scene manipulation knowledge over time for visual data, we use a dynamic knowledge graph with time attributes. A dynamic knowledge graph $G_{T_i... T_j}=(N, E)$ is a Labeled Directed Graph where edges inside are composed of the binary logical constraints in $LC$. Any edge $e \in E$ in a knowledge graph spans $LC$ and any node $n \in N$ spans $E + A$. Nodes can be spatially connected. Additionally, a timing constraint is applied in correspondence to the visual data. For any visual data within a time period $T_i... T_j$, including a clip, a video and a stream, a dynamic knowledge graph $G_{T_i...T_j}$ describes the set of relations that are presented during this time period. 

\begin{figure}[h]
\centering
    \includegraphics[scale=0.48]{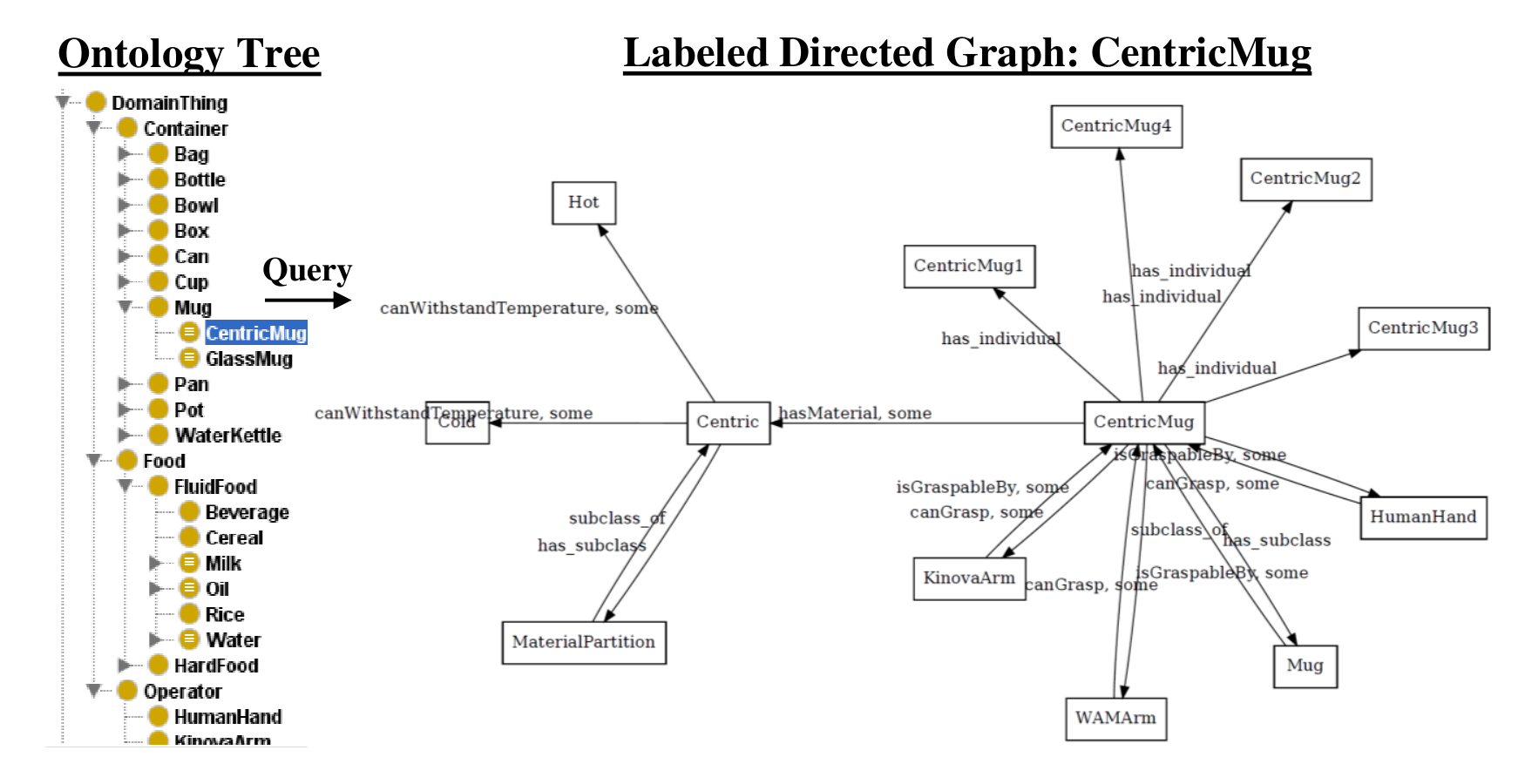}
    \caption{From ontology system to object taxonomy. An ontology system is constructed based on the entities and relations presented inside our dataset. Given a keyword concept, for example "CentricMug", we can query from the ontology and represent the associated concepts into a Labeled Directed Graph.}
     \label{fig:obj_taxonomy}
\end{figure}

For a compact inference representation, we define a command language, which can be seen as a basic skeleton form of the dynamic knowledge graph. A command language $S_{t_{i}... t_{j}} \subseteq G_{T_i... T_j}$ describes the most important relations composed by the manipulation actions at hand. A command language $S_{t_{i}...t_{j}}$ is represented as: 
$$
e_1 \xrightarrow{\text{$a_1$}} e_2 \xrightarrow{\text{$a_2$}} ...  \xrightarrow{\text{$a_n$}} e_n \eqno{(2)}
$$
\noindent where $e_1, e_2, ..., e_n\in E$ and $a_1, a_2, ..., a_n\in A$, $t_{i}... t_{j}\subseteq T_i... T_j$, and $S$ has a length of $L$. Edges inside $S_{t_{i}... t_{j}}$ are sequentially composed and restrictions can be neglected for command languages. The timing constraint also applies to the command language. For any entity $e_i$ perceived inside a command language, the commonsense knowledge over the entity can be queried from an ontology system into an Labeled Directed Graph $G_{e_i}$, which is shown in Figure \ref{fig:obj_taxonomy}. A dynamic knowledge graph $G_{T_i... T_j}$ can be seen as the union over the command language and all queried Labeled Directed Graphs: 
$$
G_{T_i... T_j} = S_{t_{i}... t_{j}} \cup G_{e_1} \cup ... \cup G_{e_n} \eqno{(3)}
$$

\subsection{Combining Vision and Language}
\begin{figure*}[ht!]
\centering
\includegraphics[scale=0.38]{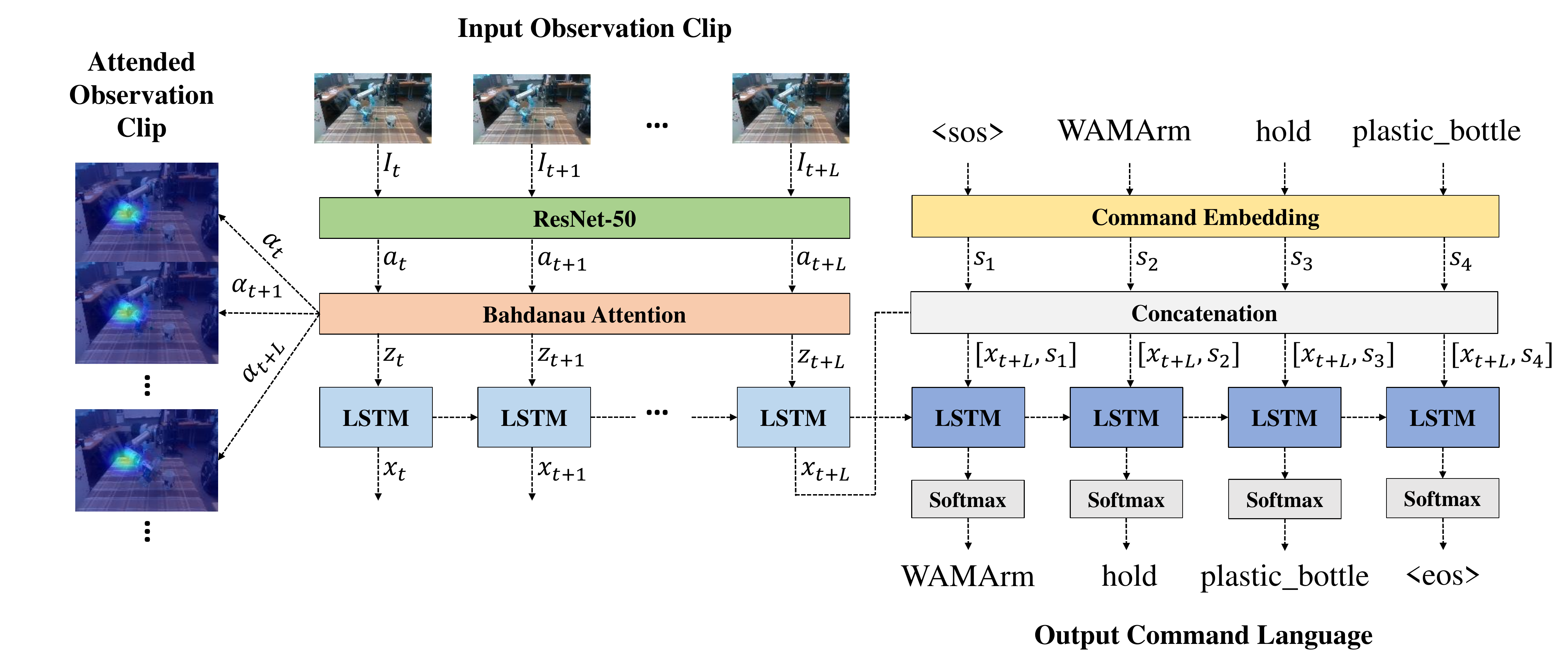}
\caption{Architecture to model command language using seq2seq with attention. An observation clip of images is first inputted into the model. Visual frame features are extracted and attended using CNN and spatial attention. An encoding LSTM encodes the sequence of attended visual features and generate an encoding vector representation, which along with the last hidden state of the encoding LSTM, is inputted into the decoding LSTM for command language generation. The model outputs the command language and the related spatial attentions over the inputted clip.}
\label{fig:seq2seq}
\end{figure*}

Given an observation as a unit of clip $C_{t_1...t_{L}} = \{I_{t_1}, I_{t_{2}}, ..., I_{t-1}, I_{t}, ..., I_{t_{L}}\}$, our goal is to caption a command language and acquire the related visual attentions over this time period $t_1...t_{L}$. We propose to train an end-to-end attention-based sequence-to-sequence (seq2seq) model to infer for a command language at any time period of the manipulation task. Figure \ref{fig:seq2seq} shows the detailed architecture for our neural command parser using seq2seq structure with spatial attention and output command language. 

\subsubsection{Spatial Attention}
Originally proposed in Xu et al. \cite{xu2015show} for image captioning tasks, the implicitly learned attention adaptively attends to relevant salient regions inside a clip of $N$ frames by the semantic labels assigned. The context vector $z_t$ at timestamp $t$ is a dynamic representation of the relevant salient part of the image feature $a_{ti}$ of size $(L, H\times W)$. A positive scalar weight $\alpha_{ti}$ is generated, interpreted as the relative importance to give to a location $i$:
\begin{align*}
e_{ti} &= f_{att}(a_{ti}, h_{t-1}) \\
\tag{4}
\alpha_{ti} &= \frac{\exp(e_{ti})}{\sum_{j=1}^{H\times W} e_{tj}}
\end{align*}
\noindent where $f_{att}$ is a mechanism that determines the amount of attention allocated to different regions of the image feature, conditioned on the previous hidden state $h_{t-1}$ of the encoding LSTM. $a_t$ can be extracted by any generic CNN network and the Bahdanau attention mechanism \cite{bahdanau2014neural} is used here. The attended visual feature is calculated simply as a weighted sum:
$$
z_t = \sum_{i=1}^{H\times W} \alpha_{ti} a_{ti} \eqno{(5)}
$$

\subsubsection{Command Language Generation with seq2seq}
Long Short-Term Memory network is a type of Recurrent Neural Network that learns long-term dependencies from the input data. Given the attended visual feature input $z_{t}$, the hidden state $h_t$ and the memory cell state $c_t$ at the next timestamp $t$ are computed as:
\begin{align*}
i_t &= \sigma (W_{ii} z_t + b_{ii} + W_{hi} h_{t-1} + b_{hi}) \\ 
f_t &= \sigma (W_{if} z_t + b_{if} + W_{hf} h_{t-1} + b_{hf}) \\
\tag{6}
g_t &= \tanh (W_{ig} z_t + b_{ig} + W_{hg} h_{t-1} + b_{hg}) \\ 
o_t &= \sigma (W_{io} z_t + b_{io} + W_{ho} h_{t-1} + b_{ho}) \\ 
c_t &= f_t * c_{t-1} + i_t * g_t \\ 
h_t &= o_t * \tanh (c_t)
\end{align*}
\noindent where $\sigma$ is the sigmoid function, $i_t$, $f_t$ and $o_t$ represent the input state, forget state and output state over the current timestamp $t$. 

The seq2seq model \cite{Sutskever2014SequenceTS} is an encoder-decoder architecture where an encoding vector representation $v$ is learned by an encoding LSTM, and a decoding LSTM learns to generate the command sentence sequence $s_1 ..., s_K$ conditioned on the encoding vector:
$$
p(s_1 ..., s_K|z_{t_1}, ..., z_{t_L}) = \displaystyle\sum_{k=1}^{K} p(s_k|v, s_1, ..., s_{k-1}) \eqno{(7)}
$$
\noindent where $v = (h_{t_L}, c_{t_L})$ is the last hidden state and the memory cell state of the encoding LSTM, $Z = (z_{t_1}, ..., z_{t_L})$ is the sequence of attended visual features and $S = (s_1, ..., s_K)$ is the corresponding output command sequence with a maximum length of $K$. $p(s_k|v, s_1, ..., s_{k-1})$ is represented with a softmax over all the tokens in the command vocabulary. The command language decoder takes the concatenation of the current command embedding feature and the last encoding hidden state as $[s_{k-1}, h_{t_L}]$ and generates the next probable command token $s_{k}$. The seq2seq structure is optimized by maximizing the log likelihood objective:
$$
\argmax_\theta \sum_{(Z,S)}\log p(S|Z; \theta) \eqno{(8)}
$$
\noindent where $\theta$ is the model parameters.

\section{Experiments}
\subsection{Implementation Details}
\subsubsection{Vision-Language Model}
The implementation for Vision-Language models are done using PyTorch. For fair comparisons, all visual features are extracted by ResNet50 pretrained on ImageNet without finetuning in all experiments. The weights for LSTMs, attention mechanism and command word embedding are randomly initialized with a hidden unit size of 512. Training is done with Adam optimizer for 100 epochs with a learning rate of 0.0001 and a batch size of 16. The maximum command sentence length is chosen as 15. 

\subsubsection{Ontology System}
The ontology system is jointly constructed using Prot{\'{e}}g{\'{e}} \cite{musen2015protege} and owlready2 \cite{lamy2017owlready}. HermiT reasoner is invoked to assess the reasoning correctness of the constructed ontology tree. There are 65 classes, 30 ID-ed individuals and 14 relations on record. 

\subsubsection{Dynamic Knowledge Graph}
In addition to an offline evaluation setting, a real-time camera stream $CS_{T_0...T_{inf}}$ is setup to observe the manipulation scene. The proposed video stream sampling method is employed to generate observation clips over time. For any observation clip $C_{t_i...t_j}$ sampled, the Vision-Language model is inferred to generate a command language $S_{t_i...t_j}$ and the related visual attentions. For each entity $e_i$ inside the command language, a word-based close matching and a recursive tree searching algorithm are jointly employed to query over the ontology tree $onto$ and and return a Labeled Directed Graph $G_{e_i}$ of taxonomy, which are further merged with the command language into the final dynamic knowledge graph $G_{T_0...T_{inf}}$. A pseudo-algorithm for generating the Dynamic Knowledge Graph is shown in Algorithm \ref{fig:alg_dkg}.
\begin{algorithm}[]
\SetAlgoLined
\textbf{Inputs:} A camera stream $CS_{T_0}$. A Vision-Language Model $Model$. A static ontology tree $onto$.\\
\KwResult{Dynamic knowledge graph $G_{T_0...T_{inf}}$ over time period $T_0...T_{inf}$.}
 initialize $CS_{T_0}$\;
 initialize an empty $G_{T_0...T_{inf}}$\;
 \While{True}{
  $C_{t_i...t_j}$ $\gets$ STREAM\_SAMPLE($CS_{T_0}$)\;
  $S_{t_i...t_j}$ $\gets$ Model($C_{t_i...t_j}$)\;
  UNION($S_{t_i...t_j}$, $G_{T_0...T_{inf}}$)\;
  \For{$e_i$ in $S_{t_i...t_j}$}{
    $G_{e_i}$ $\gets$ QUERY($onto$, $e_i$)\;
    UNION($G_{e_i}$, $G_{T_0...T_{inf}}$)
  }
 }
 \caption{Generate a Dynamic Knowledge Graph}
 \label{fig:alg_dkg}
\end{algorithm}

\begin{figure*}[!htp]
\centering
\includegraphics[scale=0.65]{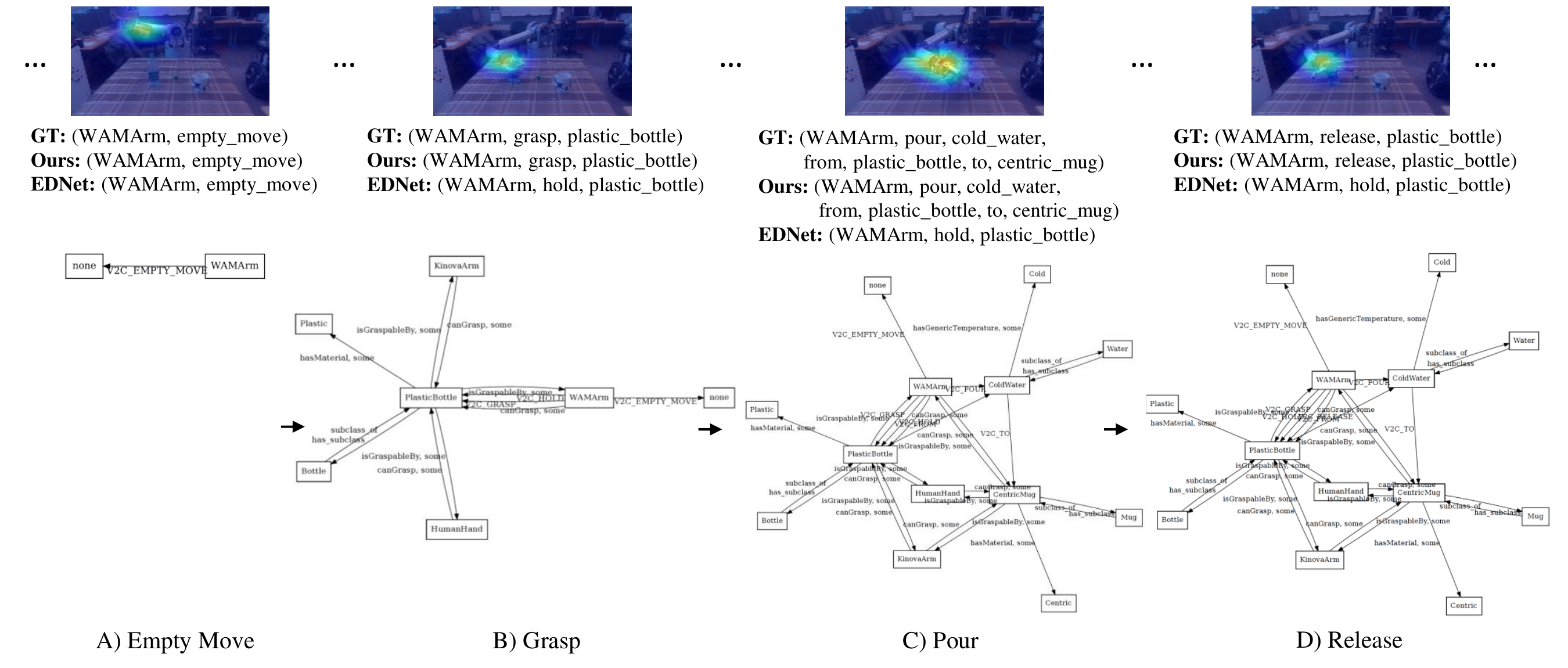}
\caption{Examples of the progressive dynamic knowledge graph visualized with spatial attentions and the command languages generated on our Robot Semantics dataset. }
\label{fig:result_kg}
\end{figure*}

\subsection{Evaluation Settings} 
\subsubsection{Datasets}
We evaluate our seq2seq architecture designs on two datasets: (1) IIT-V2C dataset, originally proposed in Nguyen et al. \cite{nguyen2018translating} to process fine-grained human action understanding in the form of command languages; and (2) the proposed Robot Semantics dataset. All experimental parameters for the IIT-V2C dataset are setup the same as in Nguyen et al. \cite{nguyen2018translating}. For experiments with our Robot Semantics dataset, three specialized evaluation divisions are combined:
\begin{itemize}

\item \textbf{Stream}: Human operators significantly hinder the smoothness of task execution by slowing down or performing a number of task-invariant motions. There are 5 human videos - 6159 images for evaluation in this category. 

\item \textbf{Unknown}: Objects that are never presented during model training are collected into this category. There are 18 human videos - 8463 images and 15 WAM videos - 22821 images in this category.

\item \textbf{Complex}: Multiple objects are presented at scene. One human uses finger to point to some objects of interests at specific locations. The manipulator performs the action on the specified objects. 7 WAM videos - 9708 are available in this category.

\end{itemize}

We use the proposed video stream sampling strategy to generate clips from videos in the Robot Semantics dataset for training and evaluation. An observation window size of 30 frames with an overlapping size of 15 frames are used, equivalent to a full 1 sec of observing under an FPS of $30$. As a result, 4188 training clips and 3075 evaluating clips are generated.

\subsubsection{Baseline Experiments}
To validate the effectiveness of our architectural design, we employ ablation studies with the following variations: 

\begin{itemize}

\item \textbf{no\_att vs. att}, where no attention is considered vs. spatial attention is employed during the stage of visual feature encoding.

\item \textbf{no\_concat vs. concat}, where only the passing of the encoding vector is employed vs. the last encoded hidden state $h_{t_L}$ is also collected and concatenated with the word embedding feature during the sequence decoding stage.

\end{itemize}


\subsection{Results and Analysis}
\subsubsection{Results for Vision-Language Model}

\begin{table}[h]
\caption{Quantitative Evaluation Results with Baselines on Robot Semantics Dataset and IIT-V2C Dataset. }
\label{result_vl}
\begin{center}
\begin{tabular}{c|c|c c c c c}
\hline
\textbf{Dataset} & \textbf{Name} & \textbf{B-4} & \textbf{C} & \textbf{M} & \textbf{R}  \\
\hline
Ours & seq2seq-concat-att & 0.581 & \textbf{0.477} & \textbf{0.784} & \textbf{4.027} \\
 & seq2seq-concat & 0.576 & 0.471 & 0.771 & 3.905  \\
 & seq2seq & \textbf{0.596} & 0.471 & 0.77 & 3.943  \\
 & EDNet\cite{nguyen2018translating} & 0.580 & 0.463 & 0.766 & 3.879  \\
\hline
IIT-V2C & S2VT\cite{venugopalan2015sequence} & 0.159 & 0.183 & 0.382 & 1.431  \\
 & SCN\cite{ramanishka2017top} & 0.190 & 0.195 & 0.399 & 1.561  \\
 & EDNet\cite{nguyen2018translating} & 0.174 & 0.193 & 0.398 & 1.550  \\
 & V2CNet\cite{nguyen2019v2cnet} & 0.199 & 0.198 & 0.408 & 1.656  \\
 & seq2seq-concat-att & 0.180 & 0.195 & 0.401 & 1.594 \\
 & seq2seq-concat & 0.203 & 0.204 & 0.417 & 1.737  \\
 & seq2seq & 0.203 & \textbf{0.209} & \textbf{0.427} & \textbf{1.765}  \\
\hline
\end{tabular}
\end{center}
\vspace{1mm}
We report the standard machine translation and language generation metrics: BLEU-4, METEOR, CIDEr, and ROUGE-L, which quantify the grammar structures and the semantic meanings of the generated sentences. All scores are computed with the coco-evaluation code \cite{chen2015microsoft}
\end{table}

Table \ref{result_vl} shows the mean over 5 experiment scores for our Robot Semantics dataset and the best experimental scores on the IIT-V2C dataset. The seq2seq-concat-att performs strongly against others, in particular, it is superior at extracting explainable visual attentions and corresponding labels of semantic meanings. This is important because human beings can attend and gaze into meaningful regions when performing manipulation tasks. Consequentially, the attention mechanism suffers when useless salient information is introduced into the video clips, as in the experiments on IIT-V2C dataset where a synthetic mean ImageNet frame needs to be padded for most video clips. Another benefit in applying attention to real-time robot manipulation is that, under our stream sampling with overlap, the manipulation concepts are consistently being attended to and traced. It can also be observed that the seq2seq structures outperform simple CNN-LSTM architectures like EDNet\cite{nguyen2018translating}, and complex Temporal Convolutional Network (TCN) based architecture like V2CNet\cite{nguyen2019v2cnet}, achieving state-of-art performance on both datasets. This indicates that the sequential modeling strategy is more viable when dealing with a real-time camera stream. Additionally, we show an example in \ref{fig:result_kg}c where EDNet fails to distinguish between the "holding" and "pouring" actions. However, with the help of spatial attention, our seq2seq-concat-att successfully captions the command language while attending to the regions of a pouring water stream. 

\subsubsection{Results for Dynamic Knowledge}
We demonstrate the dynamic evolution of the knowledge graph over time for a pouring action video in Figure \ref{fig:result_kg}\footnote{Available in our GitHub repository.}, along with the generated spatial attentions and the predicted command languages. Under the inputs from a real-time camera stream, our proposed scheme is able to dynamically visualize the evolution of a robot pouring task with the intended manipulation actions. The generated spatial attentions successfully focus on the regions where concepts of manipulation present themselves, while the manipulation procedure is summarized into the command languages. With the ontology system, command languages are completed into a constantly evolving dynamic knowledge graph filled with commonsense knowledge over presented manipulation entities. The combination of visual attention and the evolving dynamic knowledge graph fundamentally reflects the intended manipulation knowledge over the robot pouring task, which can be directly integrated into robot decision making and action execution. 

\section{Conclusions}
We present to constrain the concepts of manipulation in a taxonomic manner using an ontology system. A specific vision dataset is constructed and annotated under a strictly constrained knowledge domain for both robot and human manipulation tasks. We further propose a scheme to caption a combination of visual attentions and an evolving dynamic knowledge graph filled with commonsense knowledge. Our scheme fuses a Vision-Language model with an ontology system and works with a real-time camera stream; this enables us to interpret manipulation task visually and semantically in a timeliness manner. The experimental results show that our scheme can successfully demonstrate the possible evolution of the intended robot manipulation procedure, allowing us to prepare critical input information that can be used to decide future robotic actions. 

In future work, there are a number of things still to be explored. The stream sampling method can be augmented to include strict probabilistic modeling to consider factors such as delays. Visual attentions are also open for further interpretation, where we plan to explore the connections between language attention and human eye gaze. The visualized dynamic knowledge graph can be further expanded with graph neural networks to allow for direct integration with robotic trajectory learning during real-time action planning. Ultimately, future investigations will be focused on combining manipulation contexts with intelligent robot action controllers.





{
\bibliographystyle{IEEEtranS}
\bibliography{citation}
}

\end{document}